\documentclass[11pt]{article}

\usepackage{times}
\usepackage{graphicx}
\usepackage{amsmath}
\usepackage{amsfonts}
\usepackage{hyperref}
\usepackage{booktabs}
\usepackage{multirow}
\usepackage{geometry}
\usepackage{enumitem}
\usepackage{array} % 确保使用 p{} 列格式
\title{DINO-RotateMatch: A Rotation-Aware Deep Framework for Robust Image Matching in Large-Scale 3D Reconstruction}

\author{
	Kaichen Zhang \\ 
	Beijing National Day School, Beijing, China \\
	\texttt{kaichenzhang76@gmail.com}
	\and
	Tianxiang Sheng \\
	No.2 High School of East China Normal University, Shanghai, China \\
	\texttt{shengtianxiang0711@outlook.com}
	\and
	Xuanming Shi \\
	CodingFuture Research Center, Beijing, China \\
	\texttt{568078145@qq.com}
}

\date{}

\begin{document}

\maketitle

\begin{abstract}
	This paper presents DINO-RotateMatch, a deep-learning framework designed to address the challenges of image matching in large-scale 3D reconstruction from unstructured Internet images. The method integrates a dataset-adaptive image pairing strategy with rotation-aware keypoint extraction and matching. DINO is employed to retrieve semantically relevant image pairs in large collections, while rotation-based augmentation captures orientation-dependent local features using ALIKED and LightGlue. Experiments on the Kaggle Image Matching Challenge 2025 demonstrate consistent improvements in mean Average Accuracy (mAA), achieving a Silver Award (47th of 943 teams). The results confirm that combining self-supervised global descriptors with rotation-enhanced local matching offers a robust and scalable solution for large-scale 3D reconstruction.
\end{abstract}

\section{Introduction}

In the digital society, the proliferation of Internet technology has enabled the rapid dissemination of billions of images across diverse domains. These images represent valuable resources for applications in culture, art, and scientific research, particularly in the field of 3D reconstruction. Traditional geometry-based methods, such as Structure-from-Motion (SfM) and Multi-View Stereo (MVS), have demonstrated success in controlled environments by detecting and matching image keypoints to reconstruct detailed 3D models. Representative techniques include SIFT, SURF, and ORB, which remain fundamental to many computer vision pipelines.

Despite these advances, current approaches struggle when applied to large-scale online image collections. Internet-sourced images are often unorganized, varying significantly in illumination, viewpoint, resolution, and context—for example, drone imagery, forested scenes, or nighttime captures. These inconsistencies introduce challenges such as occlusions, viewpoint sparsity, and texture deficiencies. While deep learning–based methods, including convolutional neural networks, vision transformers, and graph neural networks, have improved robustness and contextual feature learning, they still demand substantial computational resources, lack interpretability, and rely heavily on image overlap or positional accuracy. Moreover, multi-modal fusion strategies (e.g., RGB with depth or infrared) provide incremental improvements but face barriers in standardization, generalization, and cross-domain adaptability. Consequently, a scalable and reliable solution for 3D reconstruction from unstructured online imagery is still lacking.

To overcome these limitations, this study proposes \textbf{DINO-RotateMatch}, a novel deep-learning-based framework tailored for 3D reconstruction from unorganized Internet image datasets. The contributions of this work are threefold:

\begin{itemize}
	\item \textbf{DINO-based image pairing}: Leveraging self-supervised vision transformers, our model learns general-purpose features without labeled data, achieving more efficient and accurate image pairing than exhaustive enumeration in large datasets.
	\item \textbf{Rotation in keypoint extraction}: By rotating images prior to applying ALIKED, additional orientation-dependent features are extracted, increasing the pool of potential keypoints.
	\item \textbf{Rotation in keypoint matching}: Extending the rotation strategy into the matching stage improves the number of successful correspondences and enhances robustness to viewpoint changes.
\end{itemize}

Through these innovations, DINO-RotateMatch addresses the persistent challenges of scale, diversity, and disorganization in online image collections, providing a more effective and generalizable solution for large-scale 3D reconstruction tasks.

\section{Related Work}

Image matching is a fundamental step in 3D reconstruction and has been extensively studied. Traditional approaches such as SIFT and SURF have achieved significant progress, yet they struggle with robustness under noise, illumination variation, and viewpoint changes in complex real-world scenarios \cite{lowe2004sift, bay2008surf}. In parallel, emerging deep learning–based methods offer promising alternatives, but challenges remain in terms of scalability, efficiency, and adaptability to diverse datasets.

Makantasis et al.~\cite{Makantasis2016Content} focus on accelerating 3D reconstruction from unstructured web images. By applying content-based filtering and incorporating classical feature-based techniques (SIFT/SURF, ICP), the study demonstrates the feasibility of large-scale processing. However, traditional feature matching remains sensitive to noise and viewpoint/illumination changes; ICP suffers from high complexity; and filtering strategies may mistakenly discard relevant images, leading to incomplete reconstructions \cite{besl1992icp}.

Similarly, Ntalianis et al.~\cite{ntalianis2016wild} explore cultural heritage scenarios, where retrieval and clustering serve as reordering steps combined with feature-based reconstruction. Their recognition–grouping framework successfully reconstructs several landmarks. Nevertheless, the variability of “in-the-wild’’ images reduces matching accuracy, and clustering struggles with diversity.

Johnson et al.~\cite{Johnson2022Fusion} propose a method integrating temporal cues with spatial feature extraction, improving reconstruction quality for dynamic scenes. However, real-time deployment remains challenging due to high computation costs.

Beyond geometry-driven techniques, learning-based frameworks have shown growing impact. Vision Transformer (ViT) \cite{dosovitskiy2021vit} applies Transformer architectures to patch embeddings, achieving state-of-the-art results when trained on massive datasets. Tian et al.~\cite{tian2020d2d} propose Describe-to-Detect (D2D), which reverses detect-then-describe pipelines using CNN descriptors. In a related direction, Tian et al.~\cite{8577661} introduce a differential-learning retrieval model based on gravitational-field similarity, achieving competitive results but limited scalability.

Motivated by these limitations, our research develops a multi-modal feature-enhanced 3D reconstruction algorithm combining deep features with geometric matching, further supported by intelligent image selection and filtering.

\section{Results}

In this study, we addressed the challenges of inefficiency, limited robustness, and poor scalability in 3D reconstruction from unstructured Internet images. We proposed and validated DINO-RotateMatch, a deep-learning framework designed to improve both accuracy and efficiency.
Our approach first leverages DINO to enhance image pairing efficiency. Next, we introduce rotation during keypoint extraction, enabling the model to capture additional keypoints and improve matching accuracy. Experiments demonstrate that the method significantly improves both accuracy and pairing robustness.

The dataset utilized is designed for 3D reconstruction from unstructured images. It includes a training set that contains images along with ground-truth metadata, such as dataset identifiers, file names, and pose information in the form of rotation matrices and translation vectors. This training set provides the foundation for developing our model. In addition, a set of test images is provided, covering various scenes that may include spatial overlaps or abnormal cases, which are used to debug and validate the model’s robustness. Finally, a hidden test set of approximately 1,300 images (available only for scoring notebooks) is provided to conduct the final evaluation of model accuracy.

Our model is evaluated using a single file of images that integrates multiple datasets. Without pre-assigning category labels, this setup allows us to assess the model’s ability to cluster and pair images while discarding outliers—an essential function for processing unorganized online data. Model performance is assessed in terms of image matching accuracy and clustering quality, which are jointly represented as a single score derived from two metrics. The mean Average Accuracy (mAA) serves as the primary metric, complemented by a clustering component.To compute mAA, the image clusters generated by the model are aligned with the ground-truth clusters to maximize consistency between predicted and true groupings. For each ground-truth cluster, the average accuracy—the proportion of correctly clustered images relative to the cluster size—is calculated, and the mean across all clusters yields the final mAA value.The clustering score is calculated after aligning each ground-truth scene$S_{ki}$ with one of the predicted clusters $C_{kij}$, where kdenotes the dataset index, $i$ the scene index, and $j $ the cluster index. While mAA emphasizes recall (coverage of correct images), the clustering score reflects precision (cluster purity), indicating how well incorrect images are excluded. Formally, the clustering score is defined as:

\begin{equation}
	\mathrm{CL} = 
	\frac{
		\sum_i \left| S_{ki} \cap C_{kij} \right|
	}{
		\sum_i \left| C_{kij} \right|
	},
\end{equation}

where $S_{ki}$ denotes ground-truth clusters and $C_{kij}$ the predicted clusters. The harmonic mean of mAA and CL forms the final reconstruction metric.

Finally, the harmonic mean of mAA and the clustering score is computed, analogous to the F1-score, where mAA acts as recall and the clustering score as precision. This combined metric penalizes both false positives and false negatives, thereby establishing a comprehensive evaluation framework for assessing image matching in 3D reconstruction.

\begin{figure}[]
	\centering
	\includegraphics[width=\linewidth]{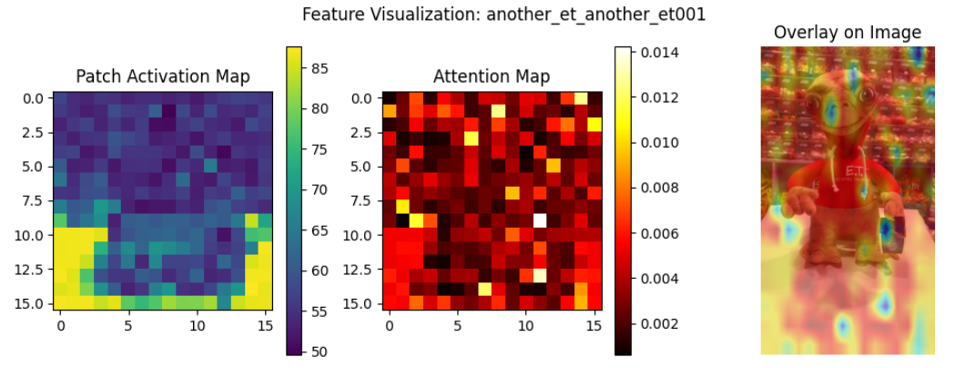}
	\caption{Feature Visualization for Image Analysis. Plots and overlay showing feature visualization results for the entity "another\_et\_another\_et001". The Patch Activation Map (left) and Attention Map (middle) depict spatial feature activation and attention distribution, respectively, with the overlay on the right showing feature weights mapped onto the original image.}
	\label{fig:4}
\end{figure}

To further investigate the effectiveness of DINO in image feature learning, we visualize the patch activations, attention distributions, and their overlay on the input image, as illustrated in Figure~\ref{fig:4} . The Patch Activation Map (left) reflects the response strength of different image patches, where warmer colors indicate stronger activations. In this example, the lower part of the image exhibits stronger activations, suggesting that the model captures basic yet distinctive features from this region. The Attention Map (middle) reveals the attention distribution derived from the self-supervised vision transformer. Unlike the relatively concentrated activation pattern, the attention is more scattered, with several localized bright regions corresponding to semantically meaningful parts of the object. Finally, the Overlay on Image (right) combines patch activation and attention, highlighting the most informative regions, such as the forehead, nose, hands, and textual patterns on the chest of the character. These areas are indeed more discriminative compared to background regions such as the table. Importantly, DINO naturally discovers such salient regions without supervision, enabling it to focus on semantically stable parts of an image across different augmented views. This property allows DINO to locate key object regions more accurately than traditional approaches, thereby facilitating more robust and discriminative representation learning, which is essential for downstream tasks such as feature extraction and image matching.

\begin{figure}[]
	\centering
	\includegraphics[width=0.5\linewidth]{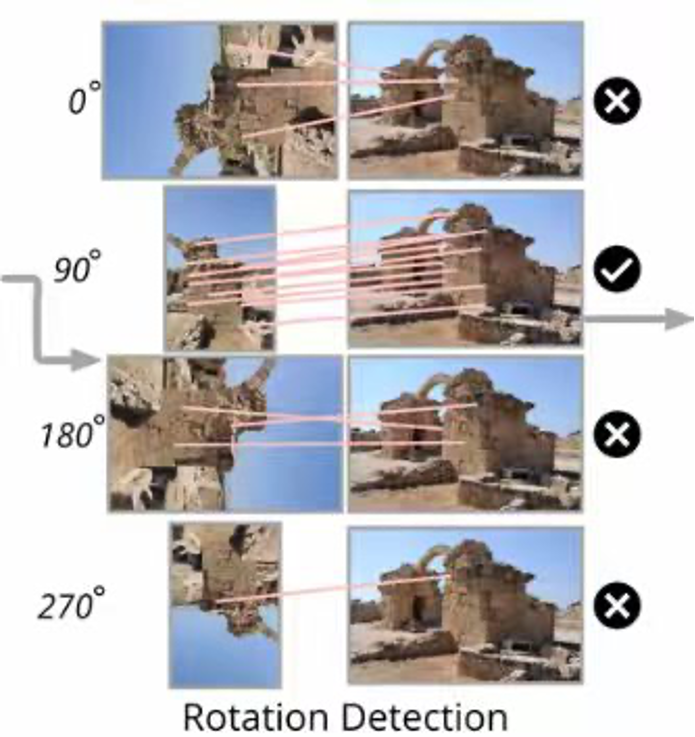}
	\caption{Rotation Detection for Image Matching. Example of rotation detection in image matching. Images of a structure at $0^\circ, 90^\circ, 180^\circ, 270^\circ$ rotations are matched, with only the 90° rotation successfully passing the matching criteria (marked with a check), while other rotations fail (marked with crosses).}
	\label{fig:5}
\end{figure}

As demonstrated in Figure~\ref{fig:5}, rotation has a significant influence on the effectiveness of feature matching. In the original orientation (0-degree rotation), only three keypoints are matched. A similar result is observed at 180°, where the number of matched keypoints also remains at three, and at 270-degree rotation the number further drops to one. In contrast, when the image is rotated by 90-degree rotation, the number of matched feature points rises sharply to ten, representing the maximum in this case. This quantitative difference demonstrates that feature matching is highly sensitive to rotation angle. Importantly, the $90^\circ$ rotation enables a substantial increase in valid correspondences, which reveals that the image pair indeed depicts the same scene and can thus be effectively used for 3D reconstruction. Therefore, the rotation strategy not only enhances the robustness of our method to changes in image orientation but also prevents potentially useful image pairs from being discarded in downstream reconstruction tasks.

\section{Discussion}

To comprehensively evaluate the effectiveness of our proposed framework, we compared DINO-RotateMatch with three representative baselines and one rotation-enhanced variant. These methods differ in image pairing strategy, keypoint extraction, and matching approach:

\begin{itemize}
	\item \textbf{Exhaustive pairing + ALIKED + LightGlue}: All image pairs are enumerated. ALIKED extracts keypoints without rotation.
	\item \textbf{DINOv2 pairing + ALIKED + LightGlue}: Image pairing relies solely on DINOv2 descriptors.
	\item \textbf{Adaptive pairing + ALIKED + LightGlue}: Exhaustive pairing for small datasets; DINO-based pairing for large ones.
	\item \textbf{Adaptive pairing + ALIKED + LightGlue + Rotation}: Extends the adaptive method by adding rotation augmentation for both extraction and matching.
\end{itemize}

We evaluated our method on both the Private and Public Leaderboards using mean Average Accuracy (mAA) as the evaluation metric. The results are summarized Table~\ref{tab:results}

First, comparing Exhaustive image pairing + ALIKE + LightGLUE with DINOv2-based image pairing + ALIKE + LightGLUE, DINOv2 consistently achieved higher scores (29.36 vs. 25.00 on the Private set, and 33.12 vs. 30.39 on the Public set). This improvement can be attributed to the self-supervised attention mechanism of DINO, which emphasizes semantically informative regions of the image and thus facilitates more accurate feature matching than the exhaustive search baseline.Second, when examining Adaptive pairing + ALIKE + LightGLUE and its rotation-enhanced variant, the rotation strategy led to further improvements. Specifically, the rotation-augmented version achieved 37.47 mAA on the Private set and 35.18 mAA on the Public set, outperforming the non-rotated counterpart (32.41 and 30.28, respectively). This indicates that incorporating rotated views during keypoint extraction and matching helps mitigate orientation-related mismatches, resulting in more robust correspondence establishment.

\begin{table}[htbp]
	\centering
	\caption{Comparison of Matching Methods on Private and Public Leaderboards (mAA)}
	\label{tab:results}
	\begin{tabular}{>{\raggedright\arraybackslash}p{6cm} >{\centering\arraybackslash}p{3cm} >{\centering\arraybackslash}p{3cm}}
		\hline
		\textbf{Matching Method Configuration} & \textbf{Evaluation Metric (mAA) - Private Leaderboard} & \textbf{Evaluation Metric (mAA) - Public Leaderboard} \\
		\hline
		Exhaustive image pairing + ALIKE + LightGLUE & 25.00 & 30.39 \\
		DINOv2-based image pairing + ALIKE + LightGLUE & 29.36 & 33.12 \\
		Adaptive pairing + ALIKE + LightGLUE & 32.41 & 30.28 \\
		Adaptive pairing + ALIKE + LightGLUE + Rotation & 37.47 & 35.18 \\
		\hline
	\end{tabular}
\end{table}

In summary, both the introduction of DINO-based attention-driven pairing and the rotation-enhanced feature extraction substantially improved performance. By combining these two strategies within our adaptive pairing framework, our method consistently outperformed the baselines on both benchmarks, demonstrating its effectiveness in achieving more discriminative and robust feature representations for image matching and 3D reconstruction.

\begin{figure}[]
	\centering
	\includegraphics[width=\linewidth]{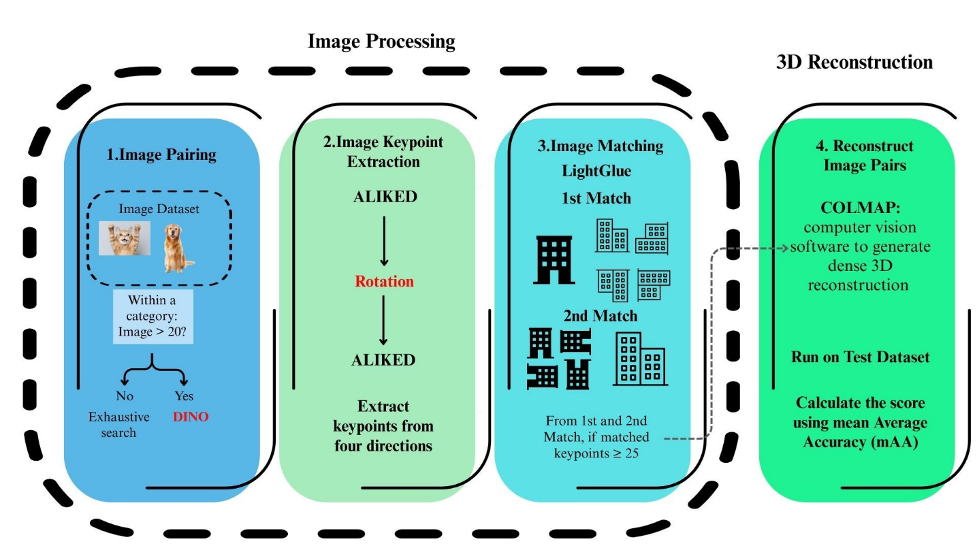}
	\caption{Overview of the DINO-RotateMatch 3D reconstruction pipeline.. Flowchart depicting the multi - stage pipeline for 3D reconstruction, encompassing image processing (image pairing, keypoint extraction, image matching) and 3D reconstruction stages. Image pairing adapts between exhaustive search and DINO - based methods based on dataset size; ALIKED extracts keypoints from four directions after rotation; LightGlue performs two - stage matching with a threshold of $\ge 25$ matched keypoints; and COLMAP generates dense 3D reconstruction, with performance scored by mean Average Accuracy (mAA) on test datasets.}
	\label{fig:1}
\end{figure}

\section{Materials and Methods}

This study introduces DINO-RotateMatch, a deep-learning-based framework designed for 3D reconstruction from unorganized Internet image datasets. The overall pipeline consists of four stages (Figure~\ref{fig:1} ): First, the raw dataset is processed through a DINO-based image pairing module, where self-supervised vision transformers are employed to learn general-purpose representations and calculate similarity between images, thereby enabling efficient and accurate pairing without exhaustive enumeration. Once paired, each image undergoes keypoint extraction using the ALIKED detector. Unlike conventional approaches, our framework incorporates a rotation strategy prior to extraction, ensuring that orientation-dependent features are captured and significantly enlarging the set of potential keypoints. The same rotation mechanism is further extended into the keypoint matching stage, where correspondences between rotated features are established, reducing mismatches and enhancing robustness against viewpoint variations. Finally, matched images and their descriptors are stored and organized into a dictionary, which is subsequently processed by COLMAP to generate 3D reconstructions. The effectiveness of the proposed framework is rigorously evaluated through mean Average Accuracy (mAA) on both validation and test datasets, demonstrating the contribution of DINO-based pairing and rotation-enhanced feature extraction and matching to the overall reliability of 3D reconstruction.

Image pairing serves as the foundation of the 3D reconstruction framework. Its goal is to efficiently identify pairs of images that share overlapping content, thereby reducing unnecessary computation in subsequent stages while maintaining high recall of valid pairs.To address datasets of different scales, we adopt a dual-path strategy (Figure~\ref{fig:2}):

\begin{figure}[t]
	\centering
	\includegraphics[width=0.95\linewidth]{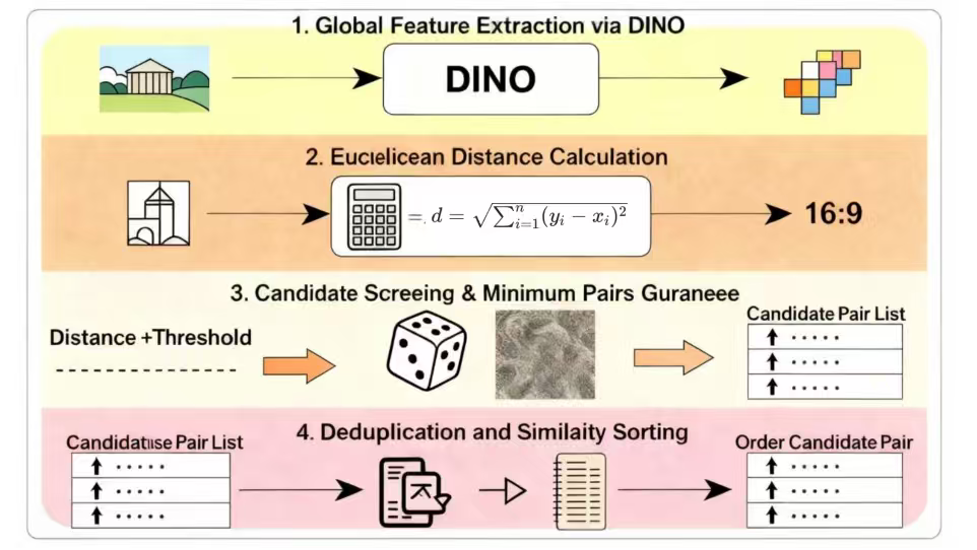}
	\caption{Pipeline of Image Feature Processing and Candidate Pair Generation. Diagram illustrating the four - stage process for image feature handling and candidate pair creation. It starts with global feature extraction via DINO, followed by Euclidean distance calculation, then candidate screening with distance thresholding to ensure minimum pairs, and finally deduplication and similarity sorting of candidate pairs.}
	\label{fig:2}
\end{figure}

\paragraph{Small-scale datasets ($N < 20$)}  
Exhaustive search is performed by generating all possible pairs$(i, j), i < j $, ensuring no valid overlap is missed.

\paragraph{Large-scale datasets ($N \ge 20$)}  
Exhaustive pairing becomes computationally prohibitive. Instead, a pre-trained DINO model extracts 768-dimensional global descriptors for each image. Pairwise Euclidean distances are computed, and the top-ranked pairs (at least 20 per dataset) are selected to ensure adequate recall, even for low-texture images.

This adaptive strategy balances accuracy and efficiency, leveraging DINO’s robustness to illumination changes and occlusions.

After image pairs are determined, keypoints and descriptors are extracted using ALIKED, a lightweight CNN-based detector with a Sparse Deformable Descriptor Head (SDDH). This design adapts to local geometry and efficiently produces descriptors with sub-pixel precision.

\begin{figure}[]
	\centering
	\includegraphics[width=0.95\linewidth]{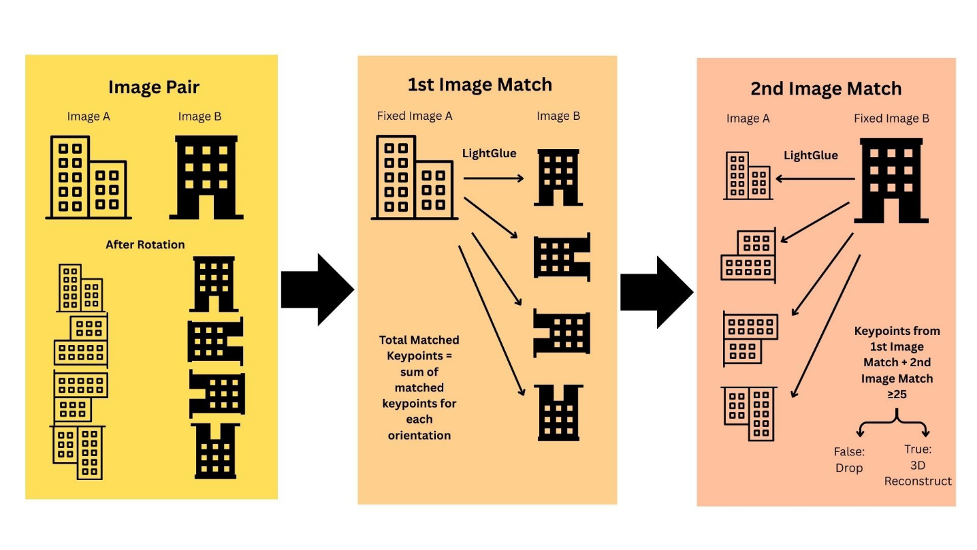}
	\caption{Image Matching and 3D Reconstruction Decision Flow. Visualization of the image matching procedure and 3D reconstruction criteria. Image pairs (Image A and Image B, shown before and after rotation) undergo two-stage matching with LightGlue. Total matched keypoints are summed across orientations in the first match. The second match fixes Image B, and if the combined matched keypoints from both stages are $\ge 25$, 3D reconstruction proceeds; otherwise, the pair is dropped.}
	\label{fig:3}
\end{figure}

To further increase keypoint diversity, each image is rotated at $0^\circ, 90^\circ, 180^\circ, 270^\circ$ before extraction. ALIKED is applied independently to each rotated version, and the resulting keypoints are aggregated. This approach ensures that features weak in one orientation may emerge in another, ultimately enhancing the robustness of subsequent matching.

Feature matching(Figure~\ref{fig:3}) is conducted using LightGlue, a transformer-based neural matcher that integrates local and global context. For each candidate pair, LightGlue establishes correspondences across all four rotated orientations. Specifically, Image A is matched against the rotated versions of Image B, and vice versa. The total number of valid correspondences is summed across both directions. If the number of matched keypoints $ \ge25$, the pair is retained for reconstruction; otherwise, it is discarded. This strategy ensures reliable correspondences while maintaining computational efficiency.

3D reconstruction is performed using COLMAP, which combines feature matching with multi-view stereo to generate dense point clouds or textured meshes. The process proceeds in two steps: (1) establish cross-image correspondences from LightGlue matches, and (2) apply multi-view stereo to densify the reconstruction.

Evaluation is conducted on datasets with ground-truth 3D models. For each reconstructed result, accuracy is measured against ground truth, and the mean Average Accuracy (mAA) is computed across all pairs. This ensures a rigorous and objective assessment of reconstruction quality.

\bibliographystyle{plain}
\bibliography{reference}

\end{document}